\documentclass[journal]{IEEEtran}
\pdfoutput=1
\usepackage{amsmath,amsfonts}
\usepackage{booktabs}
\usepackage{algorithmic}
\usepackage{subcaption}
\usepackage{array}
\usepackage{textcomp}
\usepackage{url}
\usepackage{verbatim}
\usepackage{graphicx}
\usepackage{balance}
\usepackage{hyperref}
\usepackage{color}
\usepackage{float}
\usepackage{xspace}
\usepackage{tabularx}
\usepackage{enumitem}
\setlist[itemize]{noitemsep, nolistsep, nosep}
\setlist[enumerate]{noitemsep, nolistsep, nosep}

\newcommand{\sweet}{\textit{sweet}\xspace}
\newcommand{\nvidia}{{NVIDIA}\xspace}
\newcommand{\candyds}{\tred{CandyFV}\xspace}
\newcommand{\ra}[1]{\renewcommand{\arraystretch}{#1}}
\newcommand\tred[1]{\textcolor{black}{#1}}

\begin{document}

\title{\textit{sweet}- An Open Source Modular Platform for Contactless Hand Vascular Biometric Experiments}
\author{
    \IEEEauthorblockN{David Geissbühler, Sushil Bhattacharjee, Ketan Kotwal,\\Guillaume Clivaz, and Sébastien Marcel}\\\vspace{2mm}
    \IEEEauthorblockA{\textit{Idiap Research Institute, Switzerland}}
    \thanks{This work was supported by the Innosuisse project CANDY.}
    %\thanks{Manuscript received April 19, 2023; revised August 16, 2023.\\}
}

\maketitle

\begin{abstract}
Current finger-vein or palm-vein recognition systems usually require direct contact of the subject with the apparatus. This can be problematic in environments where hygiene is of primary importance. 
In this work we present a contactless vascular biometrics sensor platform named \sweet which can be used for hand vascular biometrics studies (wrist, palm, and finger-vein) and surface features such as palmprint.
It supports several acquisition modalities such as multi-spectral Near-Infrared (NIR), RGB-color, Stereo Vision (SV) and Photometric Stereo (PS). 
\tred{Using this platform we collect a dataset consisting of the fingers, palm and wrist vascular data of 120 subjects and develop a powerful 3D pipeline for the pre-processing of this data.} 
We then present biometric experimental results, focusing on Finger-Vein Recognition (FVR). 
\tred{Finally, we discuss fusion of multiple modalities, such palm-vein combined with palm-print biometrics.} The acquisition software, parts of the hardware design, the new FV dataset, as well as source-code for our experiments are publicly available for research purposes.
%\footnote{The link to download these resources shall be shared in the paper upon publication.}
\end{abstract}


%----------------------------------------------------------
\section{\textsc{Introduction}}
\label{sec:intro}
%!TeX root=main.tex

Vascular Biometrics \cite{wang_2006}, \cite{hashimoto_2006}, or Vein Recognition (VR), offers several advantages \cite{shaheed_2018} such as convenience, high recognition accuracy, and robustness to spoofing over extrinsic biometric modalities such as face, fingerprint, or iris. 
Most existing FVR devices require the subject's finger to be in contact with the device.
They rely on transmissive Near-Infrared (NIR) illumination \cite{vanoni_2014, mohsin_2020}, where the finger is placed between the illuminator and the camera. 
The NIR light is scattered in the finger-tissue and absorbed by oxygenated hemoglobin in the blood vessels.
This design has the advantage that the sensor captures only light that has traveled through the finger and is robust to interference due to external light \cite{ramachandra_2019}. 
\begin{figure}
    \center
    \includegraphics[width=0.90\columnwidth]{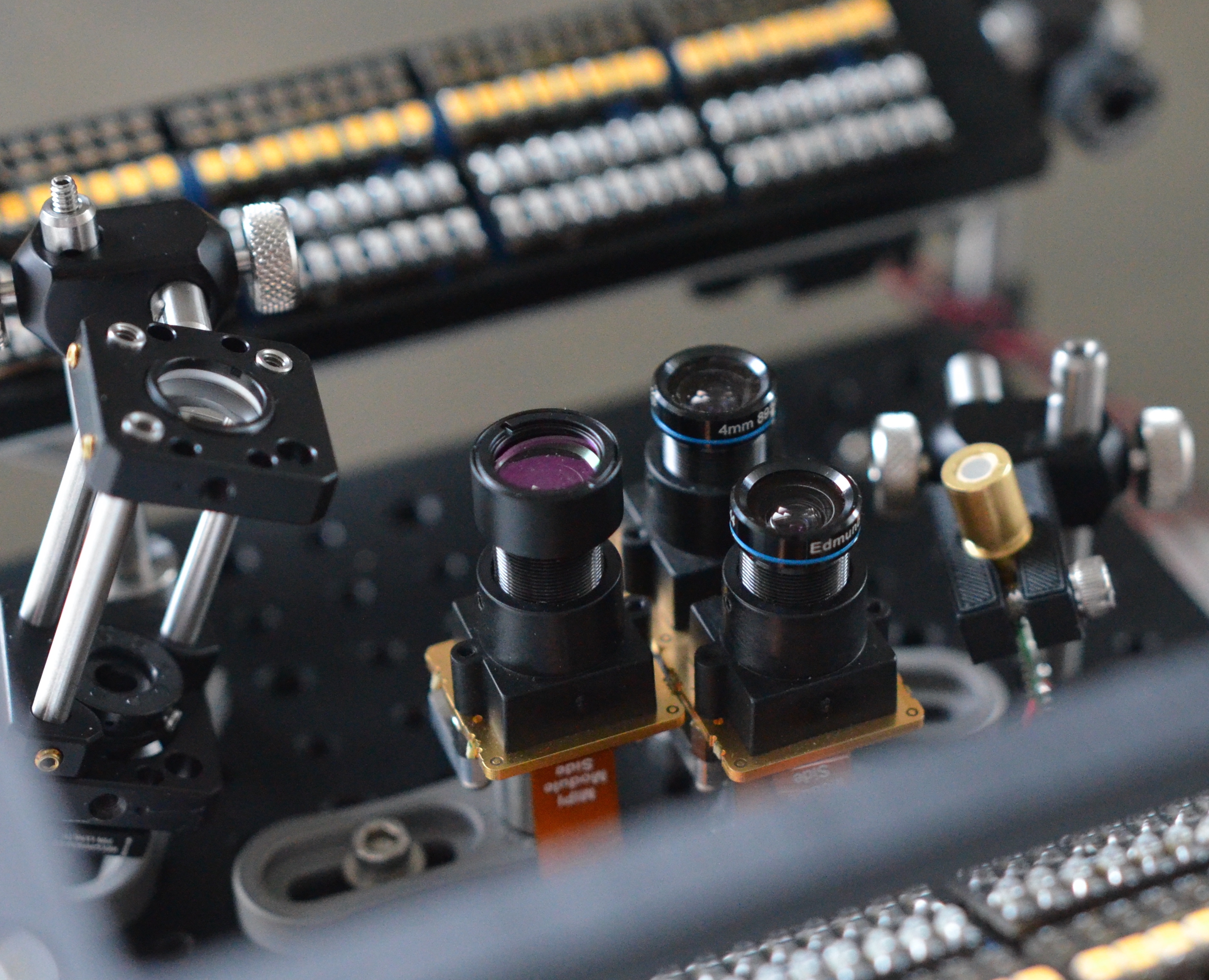}
    \caption{\small Optical bench of the \sweet sensor package.}
    \label{intro:camera}
    \vspace{-5mm}
\end{figure}

Direct contact with a biometric device can be a concern however, from the point of view of hygiene, which can be critical in environments such as hospitals where transmission of pathogens via surfaces should be mitigated \cite{otter_2013}. 
Some systems require the finger to be placed in an enclosure which users may find uncomfortable  \cite{raghavendra_2018}.

Reflection-based vascular biometric systems, on the other hand, can be made contactless \cite{michael_2010} as by their geometry the illuminator is facing the same direction as the sensor \cite{zhang_2021}. 
This family of sensors takes advantage of the penetration depth, typically a few millimeters, of NIR light into the skin tissue which enables capture of the shallow vein network. 
Unfortunately, this technique yields a signal of a much lower quality and is very sensitive to environmental conditions.

In this work we present a modular platform named \sweet aimed at exploring several technologies and sensors with the goal of improving state-of-the-art contactless reflective vascular hand biometrics. 
Our device is able to acquire images of fingers, palm and wrist in several near-infrared (NIR) wavelengths, for vascular features, as well as in visible light (RGB), for surface features such as fingerprint and palmprint. 
Moreover, this platform also captures depth information using a pair of NIR cameras with laser pattern projectors and by varying the angle of incidence of the illumination source. 
This data can be combined to reconstruct the precise shape, texture and reflectivity map, throughout various wavelengths, of the target using Stereo Vision (SV), Photometric Stereo (PS) and 3D image-alignment, yielding high quality original data unavailable, to our knowledge, in commercial hand biometric sensors.

Besides introducing the hardware prototype, we also present FVR experiments to study the efficacy of our approach.
The FVR experiments discussed here are based on a new dataset collected from 120 subjects using the \sweet platform.
While most currently available FVR sensors, including some of the most recent ones, require the user to present one finger at a time, our platform, by design, can capture vein-images of multiple fingers simultaneously by using large Field of View (FoV) optics and high resolution sensors. 
This, in turn, enables us to combine information from several fingers, to increase the FVR accuracy and reduce the FV recognition HTER to $0.057\%$. 
These results demonstrate that the \sweet platform offers state-of-the-art FVR performance.

This article is organized as follows. After a discussion about previous research related to the presented work in Section \ref{sec:rw}, we describe the hardware design for \sweet sensor platform in Section \ref{sec:hardware}. 
In Section \ref{sec:capture} we present the underlying software stack, pre-processing, calibration, SV depth inference and PS. 
FVR performance evaluated using a new dataset collected using the \sweet platform are discussed in Section \ref{sec:fvrec}.
A brief summary and conclusions in Section \ref{sec:conc} close this paper.

%----------------------------------------------------------
\section{\textsc{Related Work}}
\label{sec:rw}
%!TeX root=main.tex

Contactless vascular biometric systems \cite{michael_2010} have been investigated for more than a decade.
Reflective VR systems have been developed for finger \cite{zhang_2021}, palm \cite{tome_2015, chen_2019b}, palm and finger \cite{sierro_2015}, wrist \cite{raghavendra_2016} and forehead \cite{bhattacharya_2022}. 
Yuan and Tang \cite{yuan_2011} have also proposed combining surface and vascular features such as palm-vein with palm-print.

Multi-spectral biometric systems \cite{zhang_2016} have been used for a variety of modalities and applications.
Spinoulas \emph{et al.} \cite{spinoulas_2021} use color RGB, NIR and Short-Wave Infrared (SWIR) illumination to improve face presentation attack detection (PAD).
Multi-spectral data has been shown to improve iris recognition \cite{crihalmeanu_2012}, and fingerprint recognition \cite{rowe_2005}.
Hao \emph{et al.} \cite{hao_2008} have developed a multi-spectral imaging device for contactless palmprint recognition.

Using 3D depth information \cite{chang_2003} for face PAD is now common in consumers products such as smartphones. 
Although Stereo Vision (SV) based VR methods have been recently proposed \cite{liang_2022}, these systems are still quite rare. 
Kauba \emph{et al.} \cite{kauba_2022} propose a transmissive acquisition technique where the camera and illumination module rotates around the finger, whereas in \cite{kang_2019} three cameras surround the finger. 
In \cite{cheng_2019} photometric stereo (PS) is proposed as biometric modality using 3D knuckle patterns on the fingers.

While most apparatus in biometric laboratories are expensive research prototype, the authors in \cite{chen_2021} use a Raspberry Pi (RPi) platform for the acquisition computer and in \cite{gunawan_2018} a RPi NoiR camera is employed as a sensor.

Newly proposed FVR algorithms are compared to the state-of-the-art using publicly available datasets such as SDUMLA-HMT \cite{yin_SDUMLA11}, MMCBNU\_6000 \cite{lu13}, VERA-finger \cite{vanoni_2014}, UTFVP \cite{Ton_UTFVPDB2013}, and SCUT-SFVD \cite{qiu_2018_SCUT}.
One common characteristic of these datasets is that the biometric samples are single-finger ones.
In contrast, FV samples in the presented dataset show four fingers together, which enables finger-fusion for robust FVR.

Early research in vascular biometrics relied on various hand-crafted features such as Repeated Line-Tracking (RLT) \cite{miura_2004}, maximum curvature (MC) \cite{miura_2005}, wide-line detection (WLD) \cite{huang_2010}.
These algorithms extract binary pixel-maps representing the vein-network in the biometric sample allowing to compare them.
In this work we have used MC features.

Frequency-domain methods for FVR has also been proposed.
Yang \emph{et al}. \cite{Yang11} have used a bank of Gabor filters to enhance veins at different scales and then construct a set of FVCodes that are compared using a Cosine-similarity function.
This method is claimed to perform better than Miura's MC features \cite{miura_2005}.
These results, however, have been estimated over a proprietary, unpublished dataset.
More recently, Kovač and Marák \cite{kovac2023} have used Gabor filters to detect feature-points in vein-images.
Speeded-up robust features (SURF) \cite{bay_2008} are used to generate feature-descriptors for the selected feature-points, to construct biometric templates achieving an FVR accuracy of 99.94\% on SDUMLA-HMT.

As a recent review \cite{zhang_2022} shows, several deep learning approaches for FVR have been proposed.
Unlike for face-biometrics, publicly available FV datasets are not large enough to train a convolutional neural network (CNN) from scratch.
Up to now, deep-learning based FVR approaches have adapted pre-trained CNNs through transfer-learning on FV datasets to construct feature-extractors.
Besides actual FVR, deep-learning based methods have also been used for other purposes such as vein-segmentation, encryption, as well as
vein-enhancement~\cite{bros_BIOSIG_2021, kotwal_2022_CVPR} (see \cite{zhang_2022}).
We note here the work of Bros \emph{et al.} \cite{bros_BIOSIG_2021} proposing a  Residual Convolutional Autoencoder (RCAE) for vein-enhancement that reduces
the classification error on the UTVFP dataset from 2.1\% to 1\%.
In the present work we have used this RCAE in our processing pipeline as well.

%----------------------------------------------------------
\section{\textsc{Hardware Design}}
\label{sec:hardware}
%!TeX root=main.tex

\begin{figure*}
    \centering
    \begin{subfigure}{0.40\textwidth}
        \includegraphics[width=\textwidth]{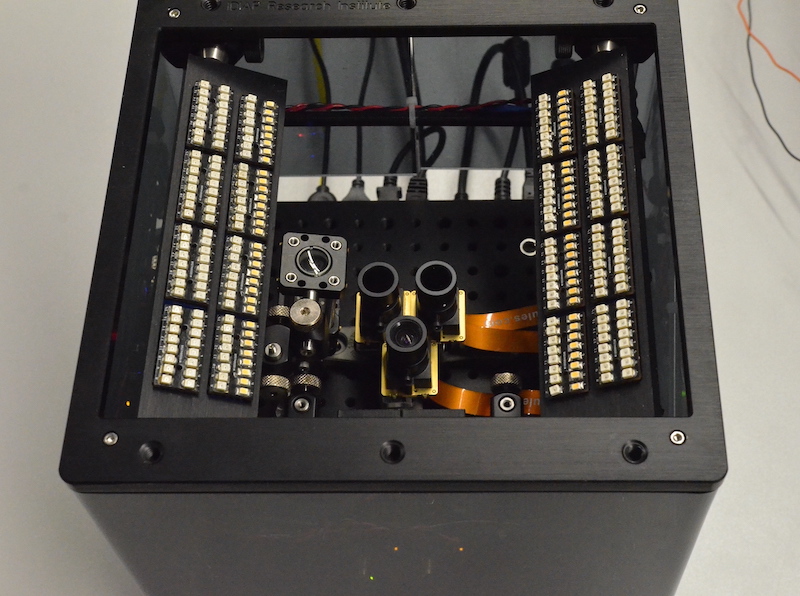}
        \caption{\sweet sensor top view.}
        \label{hard:sweet}
    \end{subfigure}
    \hfill
    \begin{subfigure}{0.57\textwidth}
        \includegraphics[width=\textwidth]{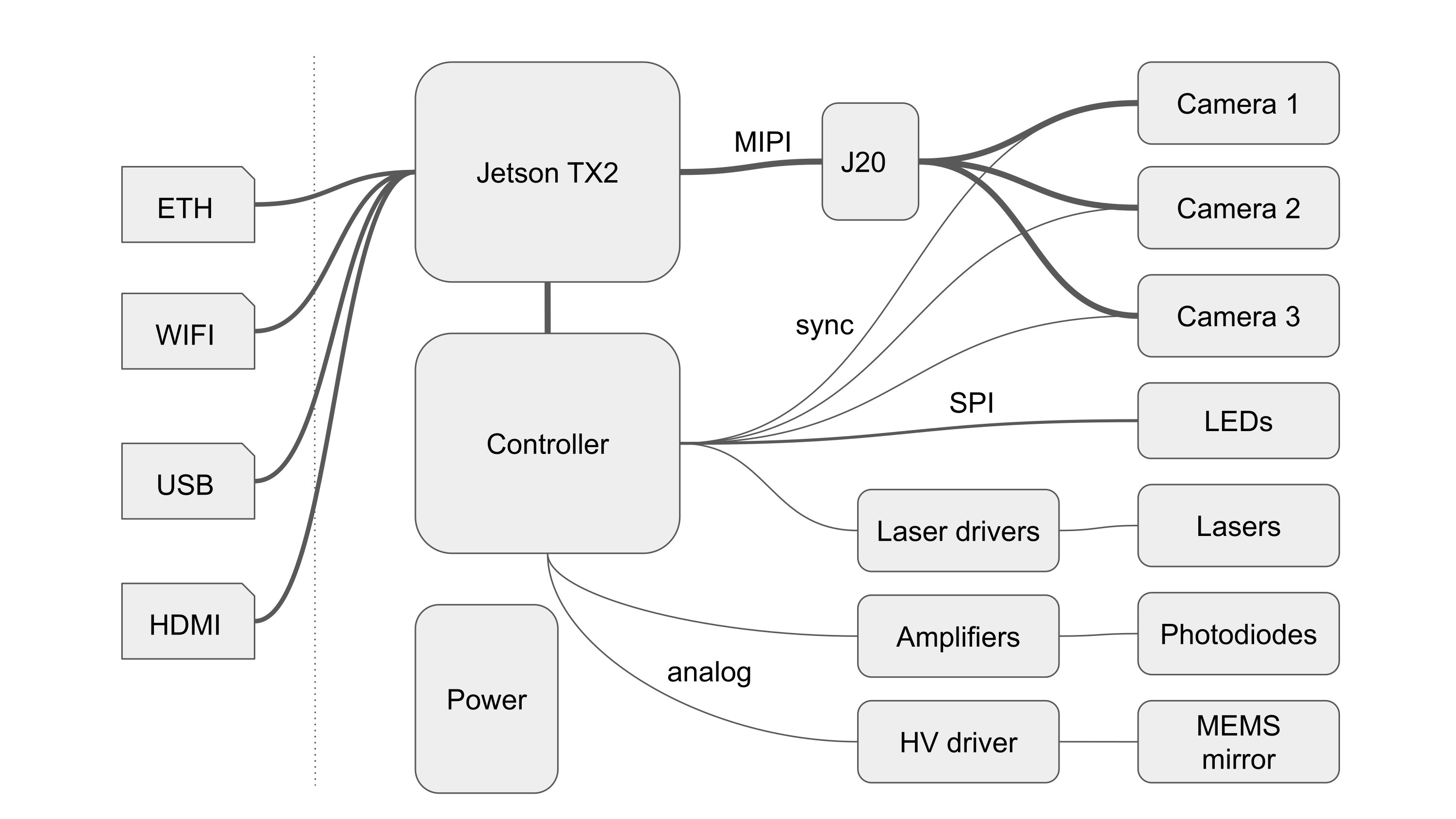}
        \caption{Connectivity diagram for the different sub systems.}
        \label{hard:diagram}
    \end{subfigure}
    \caption{The \sweet sensor platform.}
    \label{hard:platform}
\end{figure*}

\subsection{The \sweet Sensor Platform}
\label{sec:sweet}

\begin{figure*}
    \centering
    \includegraphics[width=\textwidth]{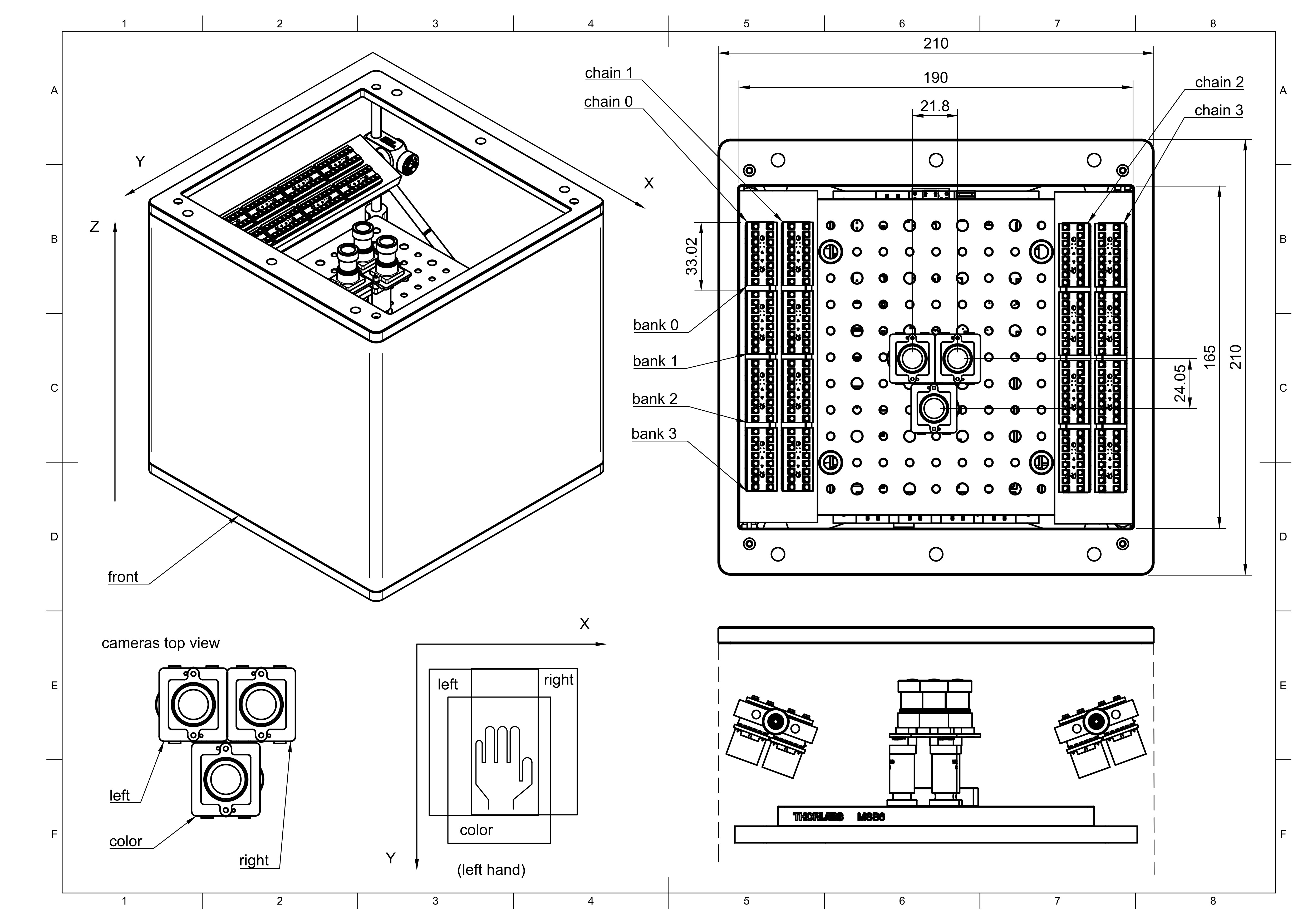}
    \caption{From left  to right, top to bottom: Coordinate system of the sensor, dimensional drawing of the sensor viewed from top, camera naming conventions and positioning, coordinate system of the captured images and cross section of the sensor showing illumination and camera positioning.}
    \label{hard:drawing}
\end{figure*}

The main goal of this work is to test several sensors and technologies that can allow contactless VR and PAD with high accuracy. In order to mitigate the difficulties associated with contactless reflective acquisition, we follow a modular and extensible approach where several cameras, sensors, illumination devices and modalities are tested for performance. We chose to use focus our attention on sensors that are sufficiently affordable for consumer products, for instance by using small camera modules and cheap S-Mount lenses. Moreover, we hope that such new sensing technologies will be useful in other domains of Science such as medical research. For the present article, we focus our attention on the following acquisition modalities:
\begin{itemize}[nolistsep]
    \item NIR HD camera pair, with multi-spectral illumination at 850nm and 950nm, for vein recognition (VR).
    \item Color HD camera with white illumination, used for surface features and PAD.
    \item Stereo Vision (SV) depth measurement using the NIR camera pair and laser dot projectors.
    \item Photometric Stereo (PS) to obtain fine grained depth resolution and texture from a set of frames illuminated from different angles.
\end{itemize}

For this platform, we made the choice of using small camera sensors and miniaturized optics to have an image quality more comparable to consumer devices but also allowing us to have a small stereoscopic baseline adapted to short range depth sensing. This choice dictates the use of small embedded computer, integrated in the platform, that can interface with the low-level MIPI-CSI data link of these camera modules. We selected a Jetson TX2 for this purpose, which also opens interesting possibilities for studying low power embedded algorithms.

A schematic diagram depicting the connection of the different subsystems of the device is shown in (Fig. \ref{hard:diagram}). The \sweet sensor platform aims at integrating the various elements, cameras, illumination, computer and electronics in a small footprint, $21\times 21 \times 21 cm$, to allow simple operation for data capture. It is completed by a screen, a keyboard and a mouse and is used like a regular computer. Cameras and optical components are mounted on an optical breadboard to allow re-positioning and extension. The enclosure (Fig. \ref{hard:drawing}) is made of two aluminum plates  connected by four $6mm$ stainless rods along which components can slide.

\subsection{Camera and Optics}

For this platform we selected a Sony IMX296 CMOS sensor board \footnote{\href{https://www.vision-components.com/en/products/oem-embedded-vision-systems/mipi-camera-modules/}{www.vision-components.com}} a sensor with good sensitivity in the NIR domain capable of capturing $950nm$ light. This sensor has a global shutter allowing for accurate synchronization required by SV. It communicates with the host via a MIPI-CSI\footnote{MIPI CSI: Mobile Industry Processor Interface - Camera Serial Interface}. It provides 1440 x 1080 pixels ($1.58M$ pixels) with 10 bits resolution, is capable of 60 Frames Per Second (FPS), and with a acceptable sensitivity in the NIR range provide it is complemented with $750nm$ low-pass filters. It has $24dB$ of analog gain and $24dB$ of digital gain, the pixels are $3.4\,\mu m \times 3.4\,\mu m$ for a total active area of $4.9 \times 3.7\,mm$, i.e. a $6.3\,mm$ diagonal or $1/2.9$ type sensor.

The sensor size constraints the focal length $f$ of the lens as the FoV should be large enough to capture an hand at a relatively short distance and wider FoV leads to higher lens distortion. We choose to use a $f= 4\,mm$ lens \footnote{\url{edmundoptics.fr/p/4mm-fl-f25-blue-series-m12-mu-videotrade-imaging-lens/31870/}}
%{Edmund Optics Blue Series M12 f=4mm f/2.5}} 
which give horizontal, vertical and diagonal angles of view of $49.64^{\circ}$, $62.97^{\circ}$ and $75.01^{\circ}$ respectively. At a working distance of $120\,mm$, the horizontal, vertical and diagonal FoV in a first order approximation is $111\,mm$, $147\,mm$ and $184\,mm$ respectively. Sensors are in portrait orientation.

The use of S-mount M12 lenses constrains the use of fixed aperture optics. On one hand a bigger aperture gives an advantage in term of collected light, on the other hand bigger apertures reduce then usable Depth of Field (DoF). This last quantity can be computed from the formulas
\begin{align}
    H &= \frac{f^2}{Nc} + f \\
    D_n &= \frac{s\left(H-f\right)}{H+s-2f} \\
    D_f &= \frac{s\left(H-f\right)}{H-s}
\end{align}
where $H$ is the hyper-focal distance, $f= 4\,mm$ is the focal length, $N = f/2.5$ the f-number of the lens, $c = 4\,\mu m$ the circle of confusion, $s$ the focussed distance and $D_n$ and $D_f$ the near and far limits. With a focussed distance $s = 120\,mm$ we get $H = 1.6\,m$, $D_n = 112\,mm$, $D_f = 129\,mm$ and
\begin{equation}
    \Delta = D_f - D_n = 17\,mm.
\end{equation}
The DoF $\Delta$ of only $17\,mm$ only correspond to an area where sharpness is maximal, the final usable range is bigger in practice. A smaller f-number of $f/5$ would lead to a double DoF ($\Delta=35.5\,mm$) range at the expense of a four time lower collected light. 

\subsection{Jetson Acquisition and Processing Platform}

The choice of MIPI CSI camera modules severely constrains the available platforms that can interface several of them. While Field Programmable Gate Array (FPGA) are ideal for these purposes, we instead chose to use an \nvidia Jetson TX2 that can acquire up to six 2-lanes camera modules. In addition of being much simpler to setup than an FPGA, these System-on-Module (SoM) are based on a multi-core ARM architecture coupled with a relatively powerful GPUs allowing the use of ML frameworks such as PyTorch. For this work a Jetson TX2 Developer Kit, with and Auvidea J20 camera expansion card, providing a dual-core \nvidia Denver 2 64-Bit CPU, 4 ARM Cortex-A57 MPCore low power cores, an \nvidia Pascal GPU with 256 CUDA cores, 8GB of shared RAM and 32GB of eMMC flash memory that is extended with a 512GB external SSD.

\subsection{Illumination, Lasers and Controller}

For LED illumination we use a modular approach first developed in \cite{spinoulas_2021} using 4 banks of 4 interchangeable modules (Fig. \ref{hard:drawing}), each with up to 16 surface mounted LEDs. Each of the 256 LED can be addressed individually in both current, up to 57mA per device, and Pulse Width Modulation (PWM). These modules are based on the PCA9745B chip\footnote{\href{https://www.nxp.com/docs/en/data-sheet/PCA9745B.pdf}{www.nxp.com}} thus requiring no LEDs resistor, are daisy chained and controlled by a serial SPI signal. The LED are powered by a dedicated and robust $5V$ DCDC Power Supply Unit (PSU) module delivering up to $8A$ to the illumination system. Switching all 256 LEDs takes approximately $1ms$. For legal reasons, we cannot release this simple design.

Since commercial laser drivers are bulky and expensive, we developed our own custom laser drivers boards based on the IC-NZN\footnote{\href{https://www.ichaus.de/upload/pdf/NZN_datasheet_D1en.pdf}{www.ichaus.de}} chip. These are capable of driving diode lasers in both optical output power (APC) or constant current (ACC) modes, switching up to $155MHz$ with an external signal, interface with most laser diodes types (P,N,M) and provide a convenient PMOD \footnote{\href{https://digilent.com/reference/_media/reference/pmod/pmod-interface-specification-1_2_0.pdf}{digilent.com}} interface. This design is released under an open source license alongside this paper.

In order to control camera triggering in real-time, LEDs dimming and lasers switching, an additional custom embedded controller board is used. This board is based on a Teensy 4.1 module\footnote{\href{https://www.pjrc.com/store/teensy41.html}{www.pjrc.com}} and will be described in a subsequent paper.

\subsection{Software}

The Jetson TX2 runs a Ubuntu 20.04 aarch64 operating system with JetPack 5.1 SDK, the kernel is patched with MIPI CSI drivers from the camera manufacturer and loads a custom device tree. The cameras frames are captured using a custom C++ Python interface that access the low-level V4L2 API \footnote{\href{https://www.kernel.org/doc/html/v4.9/media/uapi/v4l/v4l2.html}{V4L2}}. A acquisition and visualization GUI is built upon the \textit{PyQTGraph} library \footnote{\href{https://www.pyqtgraph.org/}{PyQTGraph}}, data storage and processing is based on a custom library that wraps \textit{h5py}\footnote{\href{https://www.h5py.org/}{\textit{h5py}}} and \textit{OpenCV}\footnote{\href{https://opencv.org/}{\textit{OpenCV}}}, respectively. Parts of this software suite are released under an open source license alongside this article.

\subsection{Lasers and Eye Safety Considerations}

Working with lasers, especially in a data collection environment, requires care and caution. In particular, the use of diode lasers in the $850nm$ range are a safety concern as these wavelengths can reach the cornea and there is no blinking reflex as the light is mostly invisible. We use a pair of Digigram Technology PPR-CEE850-H68V53-30k laser dots projectors \footnote{\href{https://www.digigram.com.tw/en}{Digigram}} that projects $30k$ points with a FoV of $67.7\times 53.4$ degrees. These diode lasers have a rated output power of $200mW$ and are operated at $200mA$ yielding effective optical power of $200mW$ per projector, $100mW$ after the grating. These dots projectors use a frontal grating positioned approximately $10mm$ from the diode and generate a bundle of $30k$ laser points, the power per ray is of the order of $3\mu W$ and therefore too small to cause any eye damage by itself. The center zero-order mode however is more powerful, the datasheet claims that is below $0.2\%$ of the total power, i.e. $0.4mW$. We measured a slightly higher value of $0.5mW$ for the center mode, including neighboring points, using a Thorlabs PM160 Wireless Power Meter. This implies a Class 2 standard which is not be a problem for eye safety, especially since the lasers are activated for a $3ms$ time period, five times per capture. Moreover, we implemented additional safety measures in order to protect the subjects present for data collection, by closely monitoring laser power draw, by orienting the lasers so they are not directed in the subject and operator eyes, by taking regular measurement of the optical output and by training the operator to the specifics of the system.

%----------------------------------------------------------
\section{\textsc{Calibration, Capture, Pre-Processing and Stereo Reconstruction}}
\label{sec:capture}
%!TeX root=main.tex

\newcommand{\figwidth}{0.18\textwidth}

\begin{figure*}
  %\centering
    \begin{subfigure}{\figwidth}
      \includegraphics[width=1.0\textwidth]{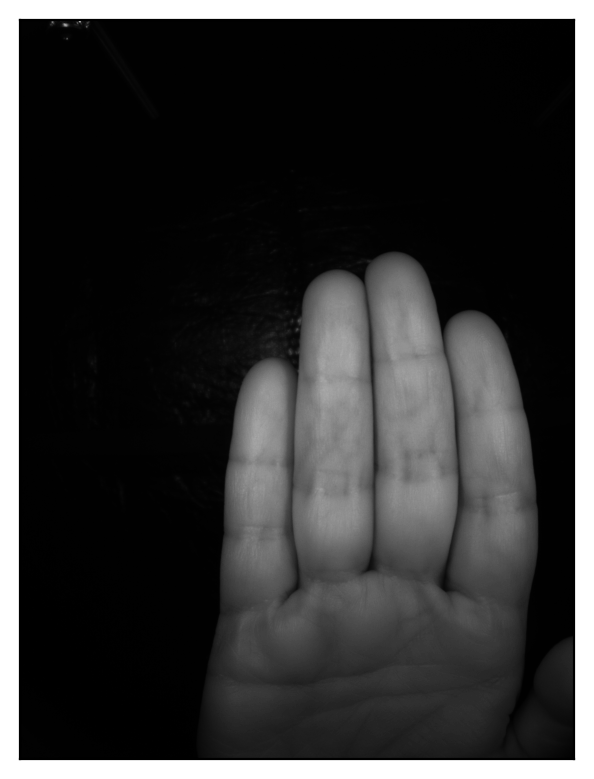}
      \caption{$850nm$}
    \end{subfigure}%
    \hfill
    \begin{subfigure}{\figwidth}
      \includegraphics[width=1.0\textwidth]{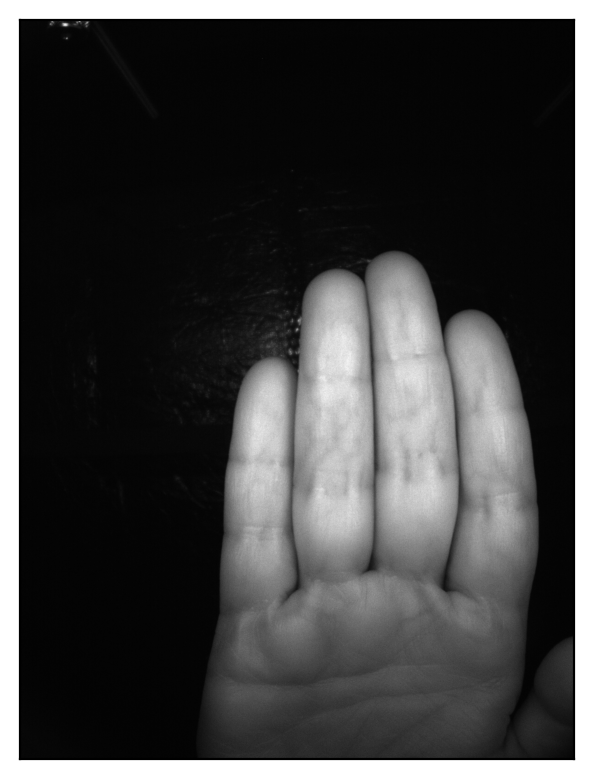}
      \caption{$950nm$}
    \end{subfigure}
    \hfill
    \begin{subfigure}{\figwidth}
      \includegraphics[width=1.0\textwidth]{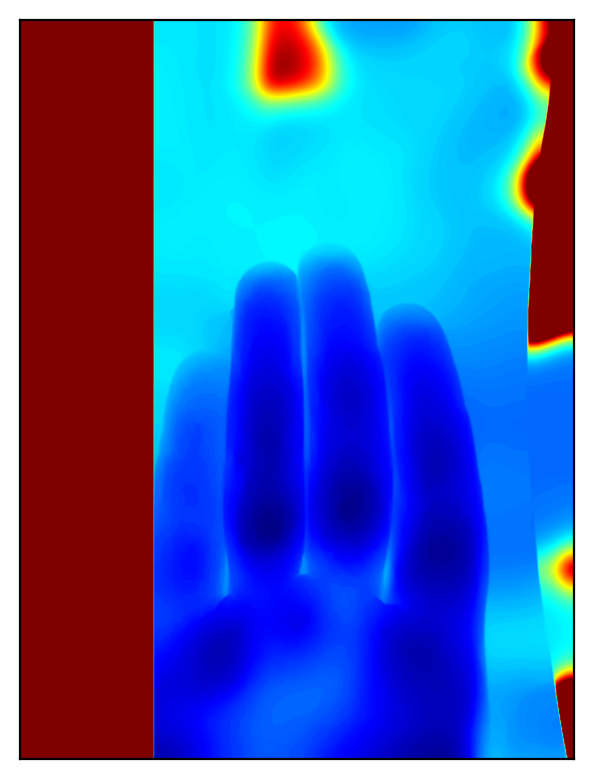}
      \caption{Stereo}
    \end{subfigure}%
    \hfill
    \begin{subfigure}{\figwidth}
      \includegraphics[width=1.0\textwidth]{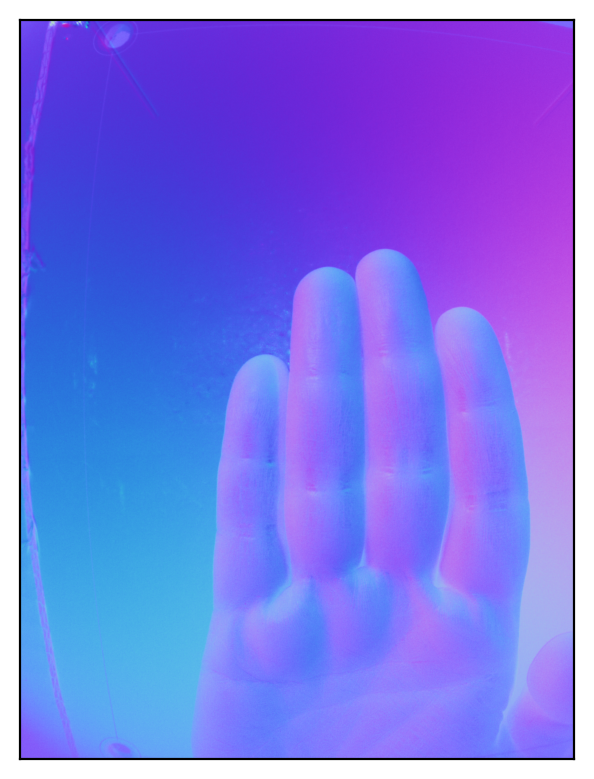}
      \caption{PS}
    \end{subfigure}
    \hfill
    \begin{subfigure}{\figwidth}
      \includegraphics[width=1.0\textwidth]{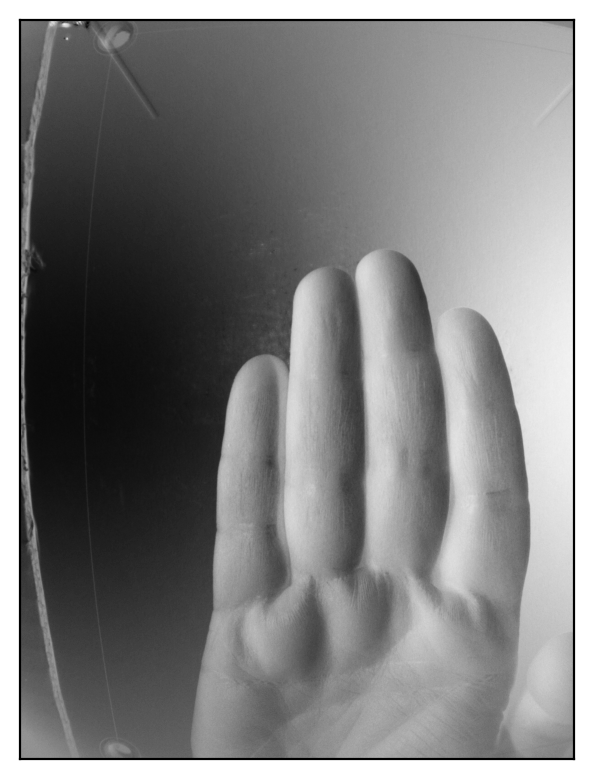}
      \caption{PS X normals}
    \end{subfigure}%
  \caption{Various channels for fingers.}
  \label{fig:fingers}
\end{figure*}

\begin{figure*}
  %\centering
    \begin{subfigure}{\figwidth}
      \includegraphics[width=1.0\textwidth]{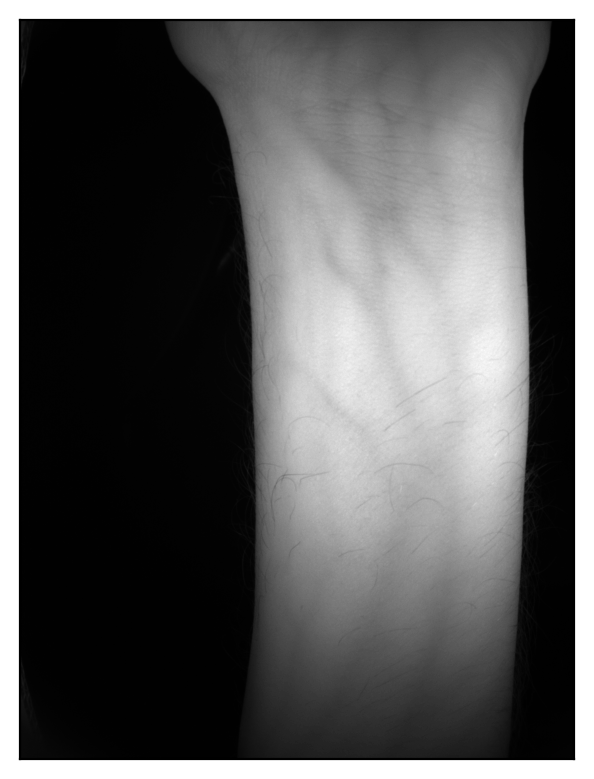}
      \caption{$850nm$}
    \end{subfigure}%
    \hfill
    \begin{subfigure}{\figwidth}
      \includegraphics[width=1.0\textwidth]{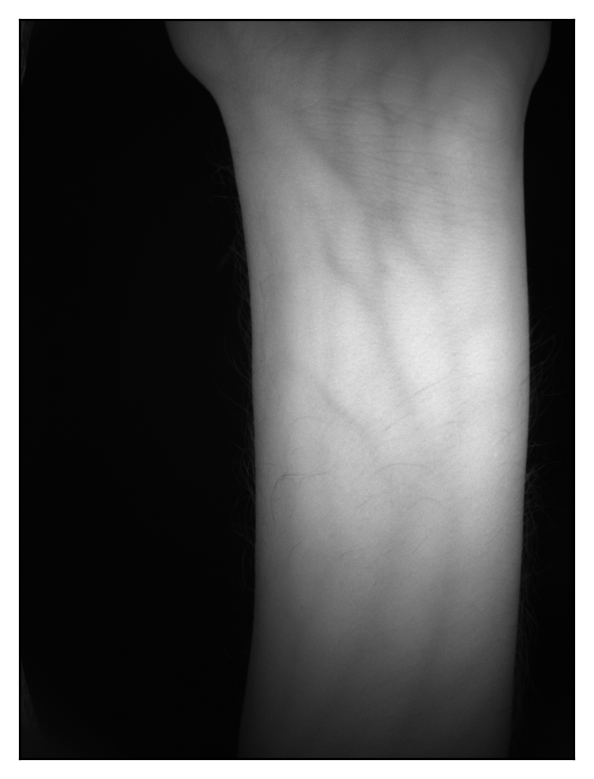}
      \caption{$950nm$}
    \end{subfigure}
    \hfill
    \begin{subfigure}{\figwidth}
      \includegraphics[width=1.0\textwidth]{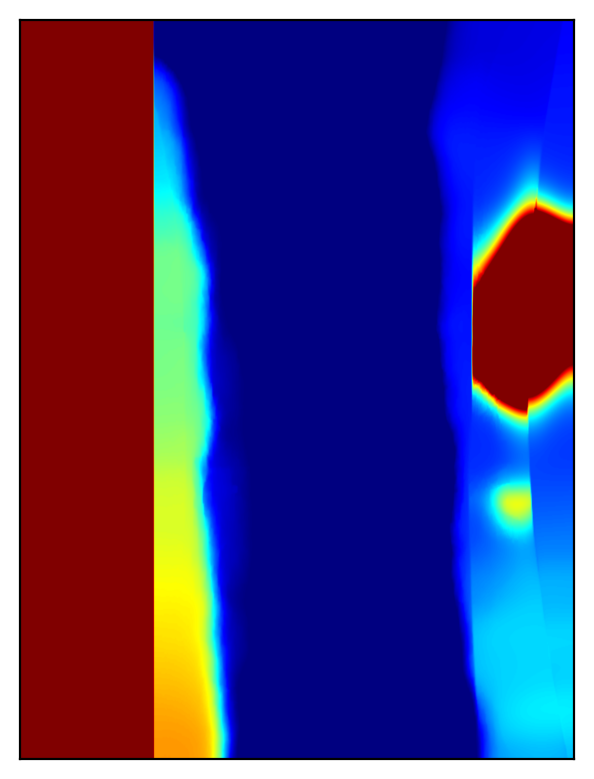}
      \caption{Stereo}
    \end{subfigure}%
    \hfill
    \begin{subfigure}{\figwidth}
      \includegraphics[width=1.0\textwidth]{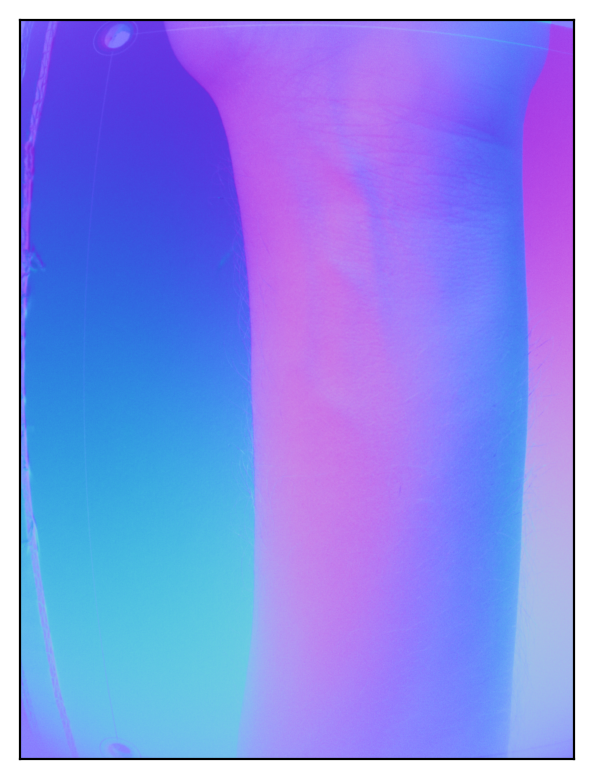}
      \caption{PS}
    \end{subfigure}
    \begin{subfigure}{\figwidth}
      \includegraphics[width=1.0\textwidth]{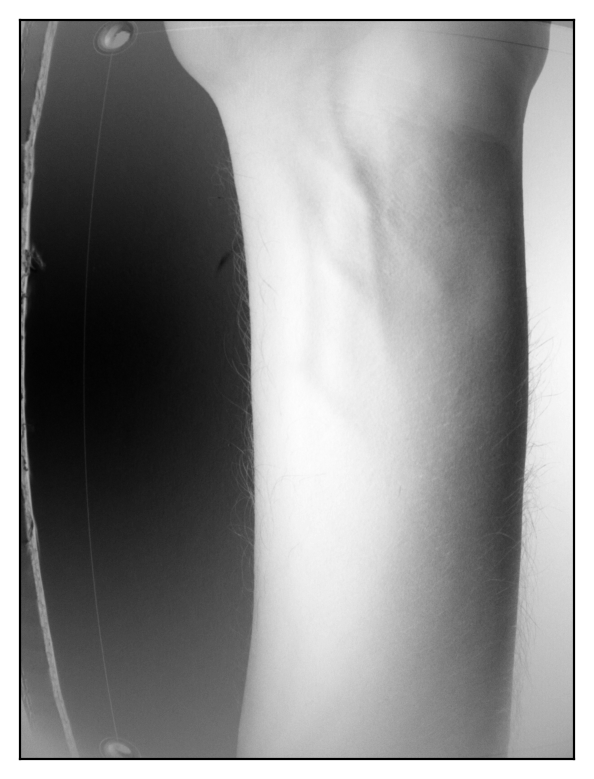}
      \caption{PS X normals}
    \end{subfigure}%
  \caption{Various channels for wrist}
  \label{fig:wrist}
\end{figure*}

In order to test which technologies perform the best, the \sweet sensor package has multiple cameras and illumination sources. 
In order to provide useful data for the algorithms, sensors, illumination and pre-processing pipelines should be precisely calibrated, a procedure described in this section.

\subsection{Illumination and Light Field Calibration}

The LEDs banks can be adjusted linearly along the $z$-axis and turned around the $y$-axis.
Moreover, since the LEDs are individually addressable so the light intensity can be adjusted along the $y$-axis as well. 
In order to find the best vertical position, angle and individual LED intensity values we use a custom a ray-tracing software to predict the light intensity and perform a manual optimization \textit{in situ} with a white paper target.

\subsection{Camera Gain Calibration and Frame Alignment}
\label{sec:gca}

When relative LEDs intensities and bank positions are set, the absolute intensity of the illumination, integration time and camera sensor gain should be setup to achieve the best possible performances. At $950nm$ both the LEDs and sensors have quite low performance which should be taken into account. The capture frame rate is fixed at 50Hz which leaves 20ms per frame to which should be subtracted the time for switching on and off the LEDs banks ($2-3ms$) as well as the sensor time to recover the image frame. In practice the maximum integration time is around $5ms$ with these parameters. In order to fix the sensor gain and the shutter time we perform a survey of noise vs gain and the optimal parameters we find is low gain ($5dB$) and medium integration time ($2ms$).

The specific camera sensors we have for this platform have a defect that a non deterministic number synchronization pulses (between 1 and 3 at $50Hz$) are missed at the beginning of the capture. We develop a method consisting of two 10 frames binary patterns we call \textit{claps} that are recorded at the beginning and at the end of the capture sequence, starting with 10 blank frames. A pre-processing routine then applies a simple pattern matching algorithm to recover the claps position, check that no frame were missed between those and trim away this data.

\subsection{Camera Calibration and Sensors Characterization}

It is essential for SV 3D reconstruction to have a very accurate characterization of the relative position of the stereo camera pair (extrinsic parameters), as well as a per camera projection matrix and lens distortion coefficients (intrinsic parameters). We perform two dozen of captures covering the whole FoV with a CharuCo target and extract the parameters with OpenCV. Intrinsic camera parameters are approximated with a 5 parameter lens distortion model and a camera matrix, while extrinsic parameters are given by a 3D rotation matrix and translation vector relative to the left camera.

\subsection{Stereo Reconstruction and Camera Views Alignment}

Reconstruction of the 3D depth map is done from a pair of rectified, un-distorted and aligned, left-right image pair. In particular, the calibration algorithm takes care that the rectification parameters, intrinsic and extrinsic, are tuned such that a point in 3D space appear at the same height in the image plane of the two cameras. If a point in 3D space appear at pixel positions $x$ on the left image and $x'$ on the right image, the difference is called the disparity
\begin{equation}
    d = x -x' = \frac{B\,f}{z},
\end{equation}
where $B$ is the baseline distance between the two cameras, $f$ is the focal length and $z$ is the distance of the 3D point to the image plane. In order to find the disparity map, and depth map, a semi-global matching (SGM) algorithm using a mutual information (Ml)-based pixel-wise matching algorithm \cite{hirschmuller_2008} implemented in OpenCV (StereoSGBM). With this depth information we can align the RGB image pixel-wise with the reference left rectified view using a simple projection algorithm.

\subsection{Photometric Stereo}
\label{sec:photometry}

We use a series of four images, taken with illumination from each corner of the LEDs banks respectively, as alternative way to extract a depth map known as Photometric Stereo (PS). It is possible to estimate the surface normals of the illuminated object using this technique proposed in \cite{woodham_1980}. The approach solves the least-square solution for the equation $I = L{\cdot}N$, with $I$ the intensities observed in the images for each pixels, $L$  the normalized vectors for each light sources and $N$ the normal surfaces. Several assumptions are assumed to simplify the problem:
\begin{itemize}[nolistsep]
  \item The light vectors are assumed constant overall the image, such as
   emitted from infinitely distant isotropic sources.
  \item The hand skin is considered with a Lambertian reflectance model, without
   specular reflection.
  \item The hand surface is assumed smooth.
\end{itemize}

In addition to this algorithm, we perform a flat-field calibration to compensate the inverse square law of light propagation, by using a flat reference with a constant albedo. This method makes the assumption that the hand is almost planar, its surface not deviating much from that of the reference plane.

%----------------------------------------------------------
\section{Vein Recognition Experiments}
\label{sec:fvrec}
%\vspace{-17mm}
%!TeX root=main.tex
%
\begin{figure}[H]
\vspace{-2mm}
\center
\includegraphics[width=0.98\columnwidth]{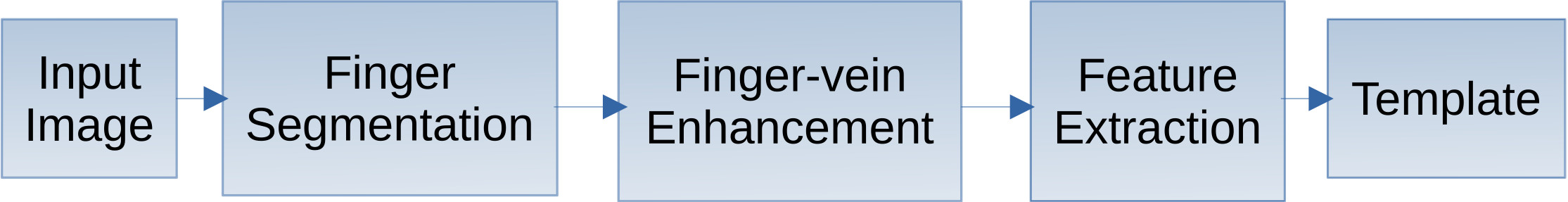}
\caption{\small Flowchart for constructing a FV-template from a FV-sample.}
\label{fig:fv_template}
\vspace{-2mm}
\end{figure}

%%%%
%
%\begin{figure}[H]
%    \begin{subfigure}{\columnwidth}
%        \includegraphics[width=0.98\columnwidth]{images/data_experiments/FV_template_fc.pdf}
%        \caption{Template extraction pipeline}
%        \label{fig:fv_seg_algo_a}
%    \end{subfigure}
%    \vfill
%    \begin{subfigure}{\columnwidth}
%    	\center
%        \includegraphics[width=0.80\columnwidth]{images/data_experiments/finger_segmentation_illus.png}
%        \caption{Finger-segmentation process.}
%        \label{fig:fv_seg_algo_b}
%    \end{subfigure}
%    \vfill
%    \caption{\small \tred{FIX Caption} The steps for finger-segmentation.}
%    \label{fig:fv_seg_algo}
%     
%\end{figure}
%%%%

A biometrics based identity-verification system functions in two phases -- \textit{enrollment} and \textit{probe}.
Both phases operate on \textit{biometric templates} -- a template being a compact representation of a biometric sample.
To enroll a new subject in the biometrics verification system, the subject provides a biometric sample.
A template, constructed from this sample, is stored in the biometrics system, associated with the subject's identity.
During the probe phase, the subject claims a certain identity, and provides a new biometric sample.
The system then compares probe-template, derived from the probe-sample, with the template enrolled for the claimed identity.
If the two templates are sufficiently similar (\textit{i.e.}, the match-score is above a predetermined threshold), we consider that the claimed identity has been verified by the system.
In this section we describe the template creation process, and then discuss the results of FV-recognition experiments with various protocols.
%\vspace{-30mm}

\vspace{-2mm}
\subsection{Finger-vein Template Creation}
%
%\vspace{-10mm}
Figure \ref{fig:fv_template} shows the flowchart of the FV-template creation process.
Each input FV-sample is an image corresponds to a presentation showing all fingers of the presented hand.
First, the four fingers -- index-, middle-, ring-, and little-finger -- are segmented out from the input image.
One template is constructed for each finger separately.

%%%
In the finger-segmentation step we process the input image to extract sub-images of individual fingers.
%The input image is expected to depict a presentation of four fingers more or less together, and the upper part of the palm.
%
%The steps for finger-segmentation procedure are illustrated in Figure \ref{fig:fv_seg_algo_b}. %\ref{fig:fin_seg_illus}.
To segment out each finger from an input image, first we generate a \textit{foreground mask} using adaptive thresholding (Otsu's method \cite{otsu_1979}) to detect the hand-region (foreground object) in the image.
Morphological opening is used to delete small regions from the resulting binary image.
% that are too small to correspond to a hand.
%
%The foreground mask is then analyzed to separate out each individual finger.
%The finger segmentation algorithm involves following steps:
%
%First, the input image is binarized using adaptive thresholding.
%selective edge enhancement to facilitate extraction of boundaries of individual fingers. For this enhancement, adaptive thresholding of input image is performed where the threshold value is calculated as a function of local average.
%
We then scan the binary image along the horizontal axis for the first foreground pixel (assumed to belong to the hand in the image).
%A finger-tip is then identified by scanning vertically (along the length of the fingers) for foreground pixels in the binarized image.% along the direction of the orientation of the fingers. The finger/ hand orientation is fixed and known to the segmentation algorithm.
%
The detected pixel corresponds to the tip of one finger (the tallest finger).
This finger is then scanned along the direction perpendicular to the vertical axis of the finger. The left and right boundaries of the finger are obtained by scanning for finger-edges on both sides.
This scanning process is repeated for each row in the image, as long as the left and right finger-boundaries extracted correspond to a reasonable finger-width.
The scanning process terminates when the estimated finger-width becomes too large for a certain row in the image.
At this point we assume that we have identified all pixels representing  a single finger.
Then we remove this finger from the image (set all finger-pixels to background) and repeat the scanning process again, this time to find another finger.
This procedure is repeated for up to four times, to detect four fingers in the image.

%\textbf{Note}: 
By the nature of the finger-segmentation process, fingers are detected in order of their height in the input image (the finger closest to the top-edge of the image is detected first, followed by the second-tallest, and so on).
In other words, the fingers are not necessarily extracted in the order index-to-little or the reverse.
In further processing, we use the relative coordinates of the center-of-gravity of each finger-mask to reorder the fingers in a natural order from index to little.
However, this finger-reordering procedure can be reliably applied only when all four fingers have been detected.
(If, for example, only three fingers have been detected, then we cannot tell whether these are index, middle and ring fingers, or middle, ring, and little fingers.)
For this reason, images where all four fingers are not detected, are excluded from further processing.
%
%The process of finger-segmentation is illustrated in Figure \ref{fig:fin_seg_illus}.
%\vspace{-1mm}
%%
%\begin{figure}[H]
%\vspace{-3mm}
%\center
%\includegraphics[width=0.90\columnwidth]{images/data_experiments/finger_segmentation_illus.png}
%\caption{\small Illustration of the finger-segmentation process.}
%\label{fig:fin_seg_illus}
%\vspace{-8mm}
%\end{figure}
%%%%

Next, a normalization step proposed by Huang \emph{et al.} \cite{huang_2010} is applied to each individual finger-image.
%The normalization step attempts to correct the orientation of the finger.
%To this end, the principal (longitudinal) axis of the finger is computed from the boundary-pixels of the finger-region, and the entire image is rotated to align the principal axis to image-boundary (vertical or horizontal, as desired).
This step simply rotates the finger-image to allign the longitudinal axis of the finger to the vertical axis as best as possible.

\begin{figure}[H]
\vspace{-2mm}
\center
\includegraphics[width=0.95\columnwidth]{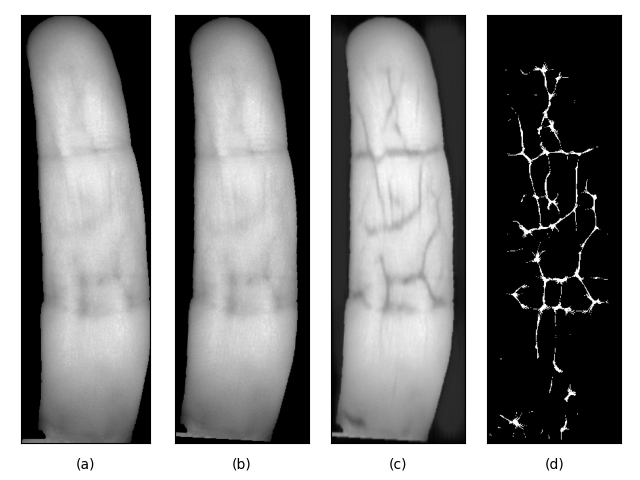}
\caption{\small Example result of the vein-enhancement. (a) Extracted finger-image; (b) normalized finger-image; (c) Vein-enhanced finger-image; (d) MC-feature-map extracted from (c).
Note the slight rotation towards the vertical axis in (b) w.r.t. (a).
%The finger-normalization step attempts to align the longitudinal axis of the extracted finger-region to the vertical axis.
The normalized finger-image, (b), forms the input to the vein-enhancement autoencoder.
}
\label{fig:vein_enh}
\vspace{-2mm}
\end{figure}

The normalized finger-image is passed to the FV-enhancement module.
Here we use a pre-trained autoencoder \cite{bros_BIOSIG_2021} to enhance the vascular structures in the input image.
Preliminary experiments showed that FV-enhancement improves the FV recognition accuracy significantly.
Hence, we have included the FV-enhancement module in our processing pipeline.
A sample result of the vein enhancement process is shown in Figure \ref{fig:vein_enh}.

FV patterns are compared based on a set of image-features extracted from the two vein-images being compared.
In this work we have used the Maximum-Curvature (MC) features \cite{miura_2005}.
%\footnote{Preliminary tests using two other feature-extraction methods, namely, wide-line detection (WLD) \cite{huang_2010} and repeated line-tracking (RLT) \cite{miura_2004} showed that the Maximum Curvature method performs much better than the WLD and RLT feature-extraction methods in our case.}.
Here, the finger-vein image is scanned line by line in the direction transverse to the length of the finger.
In each scan, the pixels of high local curvature (local second derivative of pixel-values) along the line are marked as suitable feature-points.
A sample result of the vein MC-feature-extraction process is shown in Figure \ref{fig:vein_enh}(d).
The MC feature-map extracted from an input FV-sample image is considered as the biometric template for the sample.

%\begin{figure}[H]
%\vspace{-5mm}
%\center
%\includegraphics[width=0.85\columnwidth]{images/data_experiments/fv_enh_mc_feats.png}
%\caption{\small \tred{Merge with previous figure.} Example result of the Maximum-Curvature (MC) feature-extraction step. (a) Input finger image; (b) Resulting binary feature-map extracted by the MC feature-extractor. The white pixels mark the vascular features in the input FV image.}
%\label{fig:vein_mc}
%\end{figure}

\vspace{-2mm}
\subsection{Finger-vein Matching}

%
%FV matching is a two-step process -- \textit{template} construction, followed by template-comparison.
%
%\subsubsection{Vascular Feature Comparison}
We have used the method proposed by Miura \emph{et al.}\cite{miura_2005} to compare two MC-feature based templates.
This method uses cross-correlation (computed in the frequency domain) to find the position of best match of the two input feature-maps.
The cross-correlation coefficient at the best-match position is taken as the match-score between the two templates.
In Section \ref{sec:vfr_expt} we present experimental FVR results using mainly two ISO metrics: (1) False Match Rate (FMR) and (2) False Non-Match Rate (FNMR).

\vspace{-2mm}
\subsection{Dataset Construction}
Using the \sweet platform we have collected a dataset for FV recognition experiments. 
This dataset, named \candyds,
%\footnote{The dataset shall be publicly shared for research purposes, upon publication of this paper.}
and the results of our FV recognition experiments are discussed in this section. 
\begin{table}[H]
%\vspace{-3mm}
\caption{\small Age and gender distribution of 120 subjects represented in the \candyds dataset. The first column indicates the age-groups in years.}
\ra{1.1}
\center
\vspace{-2mm}
\begin{tabular}{lcc}\toprule
\textbf{Age-Group} & \textbf{Male} & \textbf{Female}\\ \hline
18 -- 30      & 22 & 19 \\ %\hline
31 -- 50      & 20 & 18 \\ %\hline
51 and above  & 20 & 21 \\ \hline
Total         & 62 & 58 \\ \bottomrule
\end{tabular}
\label{tab:v3_dem}
\vspace{-2mm}
\end{table}
%
%Figure \ref{fig:sweet_dc} illustrates how a subject may present four fingers of a hand to the \sweet platform.
%%
%\begin{figure}[H]
%\vspace{-10mm}
%\center
%\includegraphics[width=0.85\columnwidth]{images/data_experiments/hand_over_sweet-small.jpg}
%\caption{\small Illustration of a presentation made to the \sweet platform. The horizontal metallic bar is placed only to indicate to subjects the approximate height at which to present the hand. They are informed not to rest their hand on the bar.}
%\label{fig:sweet_dc}
%\end{figure}
%
%\vspace{-05mm}
%\subsection{\textit{Bona-fide} Presentations}
\noindent
FV samples from 120 subjects comprise the \candyds{} dataset. 
For this dataset we attempted to have an even distribution of male and female subjects over three age ranges. 
The demographic distribution of the subjects who provided vein-biometrics samples for the \candyds{} dataset is shown in \hbox{Table \ref{tab:v3_dem}.}

The dataset includes five samples for each hand of each subject.
The subject presents the hand with four fingers close together (the \textit{fingers-closed} modality), over the three cameras, at a distance of roughly 10cm -- 12cm from the cameras.
%We refer to this kind of presentation as the \textit{fingers-closed} modality.
Thumbs may be entirely or partially visible in the samples; they are not used in our experiments.

Each sample includes 20 usable image-frames per camera, captured under a variety of illuminations.
In particular, each sample yields three images captured under NIR-850nm for each of the two (left, right) NIR cameras, and similarly, three images captured under NIR-950 illumination. 

%
%\begin{figure*}
%    \begin{subfigure}{0.21\textwidth}
%        \includegraphics[width=\textwidth]{images/data_experiments/Frame_01_seg_a.png}
%    \end{subfigure}
%    \hfill
%    \begin{subfigure}{0.21\textwidth}
%        \includegraphics[width=\textwidth]{images/data_experiments/Frame_01_seg_b.png}
%    \end{subfigure}
%    \hfill
%    \begin{subfigure}{0.21\textwidth}
%        \includegraphics[width=\textwidth]{images/data_experiments/Frame_01_seg_c.png}
%    \end{subfigure}
%    \hfill
%    \begin{subfigure}{0.32\textwidth}
%        \includegraphics[width=\textwidth]{images/data_experiments/Frame_01_seg_d.png}
%    \end{subfigure}
%    \vfill
%    \caption{\small \tred{FIX THIS FIG.} The steps for finger-segmentation.}
%    \label{fig:fin_seg}
%\end{figure*}

\vspace{-2mm}
\subsection{Hand Recognition Experiments}
\label{sec:vfr_expt}

We first discuss the results of several experiments for hand-identity verification based on individual fingers.
Following that we present results of hand verification based on multiple fingers, using score-fusion in various ways.

\subsubsection{Single Finger Recognition}
\label{sec:fing_rec}
%First, we discuss the performance of finger-recognition based on vascular patterns.
%As mentioned before, here the match score is simply the maximum cross-correlation coefficient of the MC feature-maps of the two finger-vein images being compared.

FV templates have been compared under eight different protocols, described in Table \ref{tab:fin_prots}.
Each protocol name is composed of three items: $<$Modality$>$\_$<$Camera$>$\_$<$NIR$>$. The Modality may be `LH' (Left hand) or `RH' (Right hand).
The Camera component `left' or `right' indicates the NIR camera from which the template has been derived, and the NIR component may be `850' or `950', indicating the illumination used to capture the image-sample.
Thus, with two options for each of three variables, we have eight different experimental protocols.
\begin{table}[H]
\caption{\small List of protocols under which finger-comparison experiments have been performed.
For each protocol, the number of enrollment images and probe images in the development (Dev) set, as well as in the evaluation (Eval) set are also listed.
}
\ra{1.1}
\center
\vspace{-3mm}
\begin{tabular}{ll|cccc}\toprule
\textbf{Id.} & \textbf{Protocol} &  \multicolumn{2}{c}{\textbf{Dev Set}}   &  \multicolumn{2}{c}{\textbf{Eval Set}}\\
~&\textbf{Name}     & \multicolumn{2}{c}{\textbf{Num. Images}} & \multicolumn{2}{c}{\textbf{Num. Images}}\\ \cline{3-4} \cline{5-6}
~ & ~                 & \textbf{Enrol.}   & \textbf{Probe} & \textbf{Enrol.}  & \textbf{Probe} \\ \hline
P1 & LH\_left\_850  & 159 & 8064 & 159 & 7836\\
P2 & LH\_left\_950  & 141 & 4908 & 138 & 4665\\
P3 & LH\_right\_850 & 159 & 7968 & 156 & 7791\\
P4 & LH\_right\_950 & 147 & 6372 & 144 & 6300\\
P5 & RH\_left\_850  & 156 & 7200 & 156 & 6998\\
P6 & RH\_left\_950  & 153 & 6138 & 136 & 4644\\
P7 & RH\_right\_850 & 156 & 7116 & 156 & 7032\\
P8 & RH\_right\_950 & 144 & 4842 & 123 & 3969\\ \bottomrule
\end{tabular}
\label{tab:fin_prots}
\vspace{-3mm}
\end{table}

For these experiments, first we partition the dataset into two subsets, named the \textit{development} (`Dev') set and the \textit{evaluation} (`Eval') set.
The Dev set is used for tuning hyper-parameters of the finger-recognition system for the desired performance.
The tuned recognition system is then applied to the data corresponding to the Eval set, to quantify the performance of the system.
The Dev and Eval sets have been constructed arbitrarily, based on the numerical subject-id assigned to each subject.
Data for the first 60 subjects
%\footnote{Subjects '0002' to '0071' have been assigned to the Dev set.} 
has been assigned to the Dev set and data for the remaining subjects has been assigned to the Eval set.

From each image captured by the \sweet platform, we extract three individual finger-vein images, corresponding to the index-, middle- and ring-finger recorded in the image.
Within each set (Dev or Eval), we have five samples for each user, for each modality.
%(Recall that, for these experiments, we have considered only two vascular-biometrics modalities -- LH and RH.)
Typically, in each sample, for each camera (left, right), we have three NIR-850nm images and three NIR-950 images.
That is, in total, for each camera we have 15 NIR-850 images per subject, and similarly 15 NIR-950 images.
Thus, for each of 60 subjects in each partition, we have 45 finger-vein images for each camera and each NIR-illumination.

The next step is to construct enrolment-sets and probe-sets.
Here, we have arbitrarily selected one sample for each modality of each subject for the enrolment sample.
The remaining samples have been designated as probe-samples.
For the finger-vein recognition experiments, each enrolled sample is considered a unique identity.

%Note that each probe-sample has not been compared against each enrolled identity.
Each probe-sample has been used for four comparisons -- one genuine comparison (with the correctly matched identity), and three \textit{zero-effort-impostor} (ZEI) comparisons (with non-matched identities).
In each ZEI comparison, the claimed-identity for a given probe template is selected randomly.
%However, the fingers are always consistent.
%\tred{In other words, for a given probe-subject, the index-finger-probe is compared only to the enrolled index-finger template of the selected ZEI-subject, and so on for the other two fingers.}
%In other words, for a given probe, the enrolled finger of the non-match subject selected for comparison is of the same kind (index-probe-sample is compared only to index-enrolled sample, and so on, for middle-finger and ring-finger).
%
For each protocol, the number of enrolled samples and probe-samples in each set (Dev or Eval) are shown in Table \ref{tab:fin_prots}.
We note two points from the Table:
\begin{enumerate}
\item the number of enrollment and probe samples for the protocols with NIR-950 illumination are consistently smaller than those for the NIR-850 protocols;
\item the numbers of images used for enrolment are not exactly 180 (three fingers for 60 subjects each).
\end{enumerate}
Both observations can be explained by the finger-segmentation results.
Images where the finger-segmentation step failed to detect exactly four fingers have been excluded from these experiments.
We noted that there were significantly more finger-segmentation errors for images captured with the NIR-950 illumination, than those captured under the NIR-850 illumination.
We estimate the finger-vein recognition rates for a specific FMR of 0.1\%.\footnote{The actual FMR achieved over the Dev set may be slightly lower than the desired FMR of 0.1\%.
This happens because, if the classification-score threshold were selected so as to allow even one additional false-match, then the FMR over the Dev set would exceed 0.1\%.}
In this analysis, the score-threshold is selected such that the FMR over the Dev set does not exceed the desired FMR.
This score-threshold is then applied the Dev set and the Eval set, to determine the actual FMR and FNMR rates over each dataset.
%Here we discuss the results for the desired FMR value of 0.1\%.\footnote{The actual FMR achieved over the Dev set may be slightly lower than the desired FMR of 0.1\%.
%This happens because, if the classification-score threshold were selected so as to allow even one additional false-match, then the FMR over the Dev set would exceed 0.1\%.}

In Table \ref{tab:fing_rec_far0.1}%
\footnote{In all the result-tables here, the FMR, FNMR and HTER values are expressed as percentages.}
we summarize the FMR and FNMR achieved for various evaluation protocols, for single finger recognition, for the FMR ceiling of 0.1\%.

Table \ref{tab:fing_rec_far0.1} shows that the FVR performance is significantly better for the right-hand fingers than for the left-hand fingers.
We do not have any logical explanation for this phenomenon.
This has probably happened for one of two reasons:
\begin{enumerate}
\item increased familiarity with the data-capture procedure -- subjects were consistently asked to present the left-hand first, therefore, presentations with the right-hand may have been `better', or
\item simply due to right-handedness of most subjects, which may lead to smaller variability in right-hand data, compared to left-hand data.
\end{enumerate}

% this table is correct
\begin{table}[H]
\caption{\small Finger-vein recognition performance at False-Match rate (FMR) of 0.1\%. 
%HTER (half-total error rate) is the average of the FMR and FNMR in each row. 
%The FMR, FNMR and HTER values indicate percentages.
The score-threshold has been computed to achieve the desired FMR (0.1\%) on the Dev set. This score-threshold is then applied to the FVR scores from the Eval set.
The lowest HTER value is highlighted in bold characters, and corresponds to the RH\_right\_850 protocol.
In general the performance for all the right-hand protocols is significantly lower than for the left-hand protocols.
}
\ra{1.1}
\center
\vspace{-3mm}
\begin{tabular}{l|rrr|rrr}\toprule
\textbf{Protocol}&\multicolumn{3}{c|}{\textbf{Dev Set}} & \multicolumn{3}{c}{\textbf{Eval Set}}\\
~ & \textbf{FMR} & \textbf{FNMR} & \textbf{HTER} 
%& \textbf{Threshold} 
& \textbf{FMR} & \textbf{FNMR} & \textbf{HTER} \\ \hline
%LFC\_left\_850 
P1 & 0.1  & 3.82 & 1.96 %& 0.089 
& 0.17 & 7.3   & 3.74 \\
%LFC\_left\_950 
P2 & 0.08 &  8.7 & 4.39 %& 0.098 
& 0.09 & 10.05 & 5.07 \\
%LFC\_right\_850
P3 & 0.08 & 5.77 & 2.93 %& 0.092 
& 0.0  & 6.46  & 3.23 \\
%LFC\_right\_950 
P4 & 0.09 & 7.74 & 3.91 %& 0.098 
& 0.02 & 9.45  & 4.74 \\
%
%RFC\_left\_850
P5 & 0.09 & 0.66 & 0.38 %& 0.088 
& 0.79 & 0.61  & 0.7 \\
%RFC\_left\_950
P6 & 0.09 & 0.57 & 0.33 %& 0.089 
& 0.66 & 0.60  & 0.63 \\
%RFC\_right\_850
P7 & 0.09 & 0.33 & 0.21 %& 0.085 
& 0.68 & 0.45  & \textbf{0.57} \\
%RFC\_right\_950
P8 & 0.09 & 1.39 & 0.74 %& 0.093 
& 0.04 & 2.27  & 1.15 \\ \bottomrule
\end{tabular}
\label{tab:fing_rec_far0.1}
\vspace{-3mm}
\end{table}

We also note that the recognition-rates achieved for protocols involving 850 nm NIR illumination are usually somewhat better than the corresponding (\textit{i.e.}, same hand, same camera) protocols involving 950 nm illumination.
This result is counterintuitive.
In theory, we expect 950 nm illumination to provide better results than 850 nm, because 950 nm NIR penetrates the soft-tissue of the fingers to a deeper extent than 850 nm NIR.
Also, 850 nm NIR tends to produce more speckle noise on the skin-surface.
On the other hand, much more power is needed for the 950 nm illumination.
Our conjecture is that the \sweet platform might not have sufficient power for the 950 nm illumination to provide the expected results.

\subsubsection{Hand-Recognition Based on Finger-Score Fusion}
In the experiments discussed so far, each finger has been considered a unique identity, that is, the identity of a subject is verified based on only a single finger at a time.
%The fundamental premise of finger-vein biometrics is that the vein-patterns in fingers are unique to every person.
%Therefore, it is tempting to assume that vein-patterns of a single finger can be reliably establish, or verify, the identity of a person.
%Can we do better by combining information from different fingers of the same hand?%\footnote{To extend this logic further, we may also combine information from the left- and right-hands of a subject to verify the subject's identity more robustly than by using only a single hand. However, we do not have sufficient number of subjects to perform `hand fusion' experiments. Therefore, we have restricted this study to individual hand-recognition only.}
%
Next, we consider each hand of a subject as a unique identity, and attempt to validate the identity of each hand based on combined recognition of three fingers of the hand -- the index-finger, the middle-finger, and the ring-finger.
The raw scores used for single-finger recognition can be combined to recognize a hand based on the vascular patterns of the three fingers combined.
For each hand, the scores the three fingers obtained in each of the eight protocols (Table \ref{tab:fin_prots}) are combined.
Thus, each hand-probe is represented by a 3-dimensional (3-D) feature-vector.
\footnote{Besides finger-fusion, we also experimented with camera-fusion, wavelength-fusion, as well as fusion of both cameras and both NIR illuminations all together, for three fingers of each hand.
However, these score-fusion strategies did not bring any further improvement to hand-recognition performance, over simple finger-fusion.
There may be two reasons for the lack of performance improvement: (1) the information captured by the two cameras, or under the two kinds of NIR illumination may not be sufficiently complementary; or (2) we probably do not have sufficient data to train the respective two-class classifiers for the camera-fusion and wavelength-fusion studies.
Therefore, due to lack of space, we have not discussed these score-fusion experiments here.}
%%The finger-score fusion may be combined in several ways.
%%
%\begin{enumerate}
%\item Multi-finger Fusion: for the hand in question, the scores the three fingers obtained in each of the eight protocols (see Table \ref{tab:fin_prots}) are combined.
%Thus, each hand-probe is represented by a 3-dimensional (3-D) feature-vector.
%%
%\item Camera Fusion: the scores of the three fingers obtained for the left camera and for the right camera are combined to construct a 6-D feature-vector to represent the hand.
%%
%\item Wavelength Fusion: the scores of the three fingers obtained for each NIR illumination (850 nm and 950 nm), using the same camera (left or right), are combined to construct a 6-D feature-vector to represent the hand.
%%
%%\item Camera and Wavelength Fusion: the scores of the three fingers obtained for each NIR illumination and for each camera, are combined to construct a 12-D feature-vector to represent the hand.
%%
%\end{enumerate}

%These score-fusion experiments to identify individual hands are discussed below.
%In these score-fusion experiments we consider each hand of each subject as a unique identity.

The general approach to score-fusion adopted in this study is as follows:
feature-vectors are constructed for each hand-identity using the individual finger-comparison scores.
While constructing these feature-vectors, finger-scores are selected either only from genuine-probes, or only from ZEI-probes.
In this way we obtain, for each hand-identity, a set of genuine-probe (`match') feature-vectors, and another set of ZEI (`non-match') feature-vectors.
A two-class classifier is then constructed using the feature-vectors in the Dev set.
This classifier is used to label the feature-vectors of the Eval set, to quantify the hand-recognition performance of the system.
In this study, we have used Support Vector Machines (SVM) with RBF (radial basis function) kernels, for the score-fusion experiments.

%\paragraph{Multi-Finger Fusion}
%\label{sec:mff}
In each protocol in Table \ref{tab:fin_prots}, we fuse the FVR-scores of the three fingers of the hand corresponding to the protocol.
% (\textit{e.g.}, left-hand identities in the LH\_left\_850 protocol).
Thus, all probe feature-vectors used in a given experiment represent information from the same hand, captured by the same NIR camera, under the same NIR illumination.
The results of FVR-score fusion within each FVR protocol are shown in Table \ref{tab:fing_fusn_rec_far0.1}.
These numbers quantify the performance of the score-fusion system when the FMR over the Dev set is limited to 0.1\%.
That is, the classification-score threshold\footnote{Note that, for the fusion experiments, the score-thresholds are not the raw finger-scores, but the scores generated by the SVM classifier for each probe.
To avoid confusion, in these fusion experiments we refer to the threshold applied to the output of the 2-class classifier as the `classification-score-threshold'.}
is selected in such a way that the FMR over the Dev set does not exceed 0.1\%.
This classification-score-threshold is then applied to the classification-scores generated for the Eval set, to estimate the FMR and FNMR over the Eval set.
\begin{table}[H]
\caption{\small Results of score-fusion of three fingers of a hand within the same protocol. The best performance, printed in bold, corresponds to the protocol `RH\_right\_850'.
%\tred{The recognition rates shown here have been achieved by setting a limit of 0.1\% on the FMR of the Dev set.}%
}
\ra{1.1}
\center
\vspace{-3mm}
\begin{tabular}{l|rrr|rrr%|p{1.6cm}
} 
\toprule
\textbf{Prot-}&\multicolumn{3}{c|}{\textbf{Dev Set}} & \multicolumn{3}{c}{\textbf{Eval Set}} %&\textbf{Number}
\\
\textbf{ocol} & \textbf{FMR} & \textbf{FNMR} & \textbf{HTER} 
%& \textbf{Threshold} 
& \textbf{FMR} & \textbf{FNMR} & \textbf{HTER} %&  \textbf{~~of SV}
\\ \hline
%LFC\_left\_850
P1 & 0.1  & 2.68 & 1.39 %& -0.539 
& 0.20 & 4.29  & 2.25 %& 76 of 2688
\\
%LFC\_left\_950
P2 & 0.08 & 4.2 & 2.14 %& -0.746 
& 0.54 & 3.57 & 2.06 %& 78 of 1636
\\
%LFC\_right\_850
P3 & 0.05 & 2.71 & 1.38 %& -0.519 
& 0.05  & 3.93  & 1.99 %& 86 of 2656
\\
%LFC\_right\_950
P4 & 0.06 & 3.04 & 1.55 %& -0.747 
& 0.6  & 2.67 & 1.66 %& 74 of 2124
\\
%
%RFC\_left\_850
P5 & 0.06 & 0.0 & 0.03 %& -0.903 
& 0.46 & 0.0  & 0.23 %& 20 of 2400
\\
%RFC\_left\_950
P6 & 0.07 & 0.0  & 0.03 %& -0.918 
& 0.45 & 0.0  & 0.23 %& 17 of 2046
\\
%RFC\_right\_850
P7 & 0.06 & 0.0 & 0.03 %& -0.862 
& 0.11 & 0.0  & \textbf{0.06} %& 22 of 2372
\\
%RFC\_right\_950
P8 & 0.08 & 0.46 & 0.27 %& -0.898   
& 0.0  & 1.51  & 0.76 %& 19 of 1614 
\\ \bottomrule
\end{tabular}
\label{tab:fing_fusn_rec_far0.1}
\vspace{-3mm}
\end{table}

First, we note that FVR-score fusion improves the hand-recognition performancen compared to single finger FVR.
%Consider the scenario where hand-recognition is based on any single finger.
For the left hand, the single FVR error-rates (HTER in Table \ref{tab:fing_rec_far0.1}), %\ref{tab:midl_fing_rec_far0.1} and \ref{tab:ring_fing_rec_far0.1}) 
range from 3.5\% to 5\% for each of the individual fingers.
Finger-fusion reduces the left-hand recognition error-rates to about 2\% or lower in all four left-hand protocols.
% 
%Given that in the single-finger recognition experiments the right-hand protocols showed significantly better performance than the left-hand, 
%It is no surprise that in the finger-fusion experiments also,
%Again, the results for hand-recognition for the right-hand protocols are clearly better than for the left-hand protocols.
For the right-hand, single-finger FVR performance is already very high (FVR in protocols P5 -- P8 in  Table \ref{tab:fing_rec_far0.1}).
Multi-finger FVR performance for the right hand still reduces the classification error.
The best performance, a HTER of 0.06\% for the `RH\_right\_850', is a 10-fold improvement over single-finger FVR in the same protocol.% (compare with Table \ref{tab:fing_rec_far0.1}).

\section{\textsc{Concluding Remarks}}
\label{sec:conc}
%!TeX root=main.tex

In this paper we have described the design of a contactless vascular biometrics platform named \sweet, that has been developed for the purpose of testing of various technologies and methods for contactless vascular biometrics.
With this platform we can record hand vascular images in NIR wavelengths (850 and 950 nm)
and skin surface in in color RGB.
With a pair of NIR cameras and by varying illumination incidence angle we can also extract precise and detailed depth and surface-normal maps using SV and PS, respectively.

In this work we have focussed on finger-vein (FV) biometrics using this platform. 
Unlike existing FV sensors that capture vascular data from only one finger at a time, the \sweet platform is designed to image four fingers of a hand simultaneously.
This enables us to implement multi-finger vein recognition, which is not possible using previous FV devices.

We have collected a large dataset of vascular biometrics samples from 120 subjects.
The FV data collected from these subjects has been curated into a dataset named \candyds.
In this paper we report FV recognition results based on this dataset.

We start with FV recognition experiments where each finger of a subject is treated as a unique identity.
Only images of the index-, middle-, and ring-finger of each hand have been included in these experiments.
For each finger we have eight test protocols -- all possible combinations of two hands, two cameras, and illumination using two NIR-wavelengths.
A probe-sample is compared to an enrolled reference using the Miura-Match method based on maximum-curvature (MC) vein features \cite{miura_2005}.
In all experiments the finger-veins have first been enhanced using an autoencoder based method \cite{bros_BIOSIG_2021}.

We found that the data for the right hand provides better FV recognition results than the left hand.
In particular, the best single finger recognition result (HTER of 0.6\%) has been achieved for the right hand using data captured by the right NIR camera under 850 nm illumination.
(See Table \ref{tab:fing_rec_far0.1}.)
We do not see any systematic reason why the FVR results for the right hand should be better than those for the left hand.
Perhaps, because most subjects are right handed, the data recorded for the right hands of subjects may have smaller variability.

Next we have explored score-fusion, to combine information from multiple fingers, to identify a hand.
In these score-fusion experiments, each hand is considered as a unique identity, represented by three fingers.
The FVR scores for the three fingers the hand are used to construct a feature-vector representing the hand.
A two-class support vector machine (SVM) classifier is then trained to verify the identity of the presented hand.

Fusing the scores of the three fingers within each protocol (same camera, same wavelength), we can achieve a hand-recognition error rate of 0.06\% (a 10-fold improvement over the error-rate of recognition based on only one finger).
%Multi-camera score fusion and multi-wavelength score fusion did not improve upon the simple finger-fusion result.

Our platform has been designed in a modular fashion and allows for a large range of improvements and extensions. In the future we will extend this platform in several ways to improve its acquisition performances, for instance by adding side facing cameras, adding sensors with better performance in the NIR range and by experimenting with illumination, for example using polarized light. We also want to test other technologies, such as Single Pixel Laser Detectors (SPLD) to add capabilities in the SWIR range. We will also improve the pre-processing pipeline with channel fusion such as SV with PS fusion.
Moreover, the fast acquisition frame-rate facilitates even more complex algorithms based on image sequence analysis, which we will explore in the future.

%In the CANDY project we have focussed on developing the \sweet platform, and estimating its efficacy in finger-vein based person identification.
%We have collected a large and diverse set of vascular biometrics data using this platform.
The \candyds dataset offers a large scope for further research.\\
\textit{Fusion}: Up to now we have only explored score-based fusion.
This strategy worked well for combining multiple fingers, but did not lead to further performance-improvement when applied to camera fusion or NIR-wavelength fusion.
Next we aim to study data-fusion and feature-fusion methods for combining information from the two NIR cameras, as well as from the two NIR illumination wavelengths.\\
\textit{End-to-end processing}: Our initial experiments with CNN based end-to-end FV recognition have not shown results comparable to those presented in this report.
This is why end-to-end FV recognition has not been discussed here.
In future work we plan to explore this direction more rigorously.\\
%\textit{PAD}: We plan to further refine the surface-normal based PAD method proposed here. We also plan to study ways of combining the shape information encapsulated in the surface-normal maps with other cues to improve PAD for a variety of PAI species.\\
%\textit{Palm-Vein and Wrist-Vein Recognition}: So far we have only worked with the finger-vein data collected in this project.
%We have also recorded palm-vein and wrist-vein data in this project.
%This data will support similar experiments for contactless palm-vein recognition and wrist-vein recognition, as well as fusion of several vascular modalities.

%----------------------------------------------------------
\vspace{-5mm}
\section*{Acknowledgement}
\vspace{-1mm}
This work has been supported by the Innosuisse project CANDY, in collaboration with GlobalID. The authors also gratefully acknowledge the contributions of Ms. Karine Vaucher, Mr. Victor Bros, Mr. Jérémy Maceiras, Mr. Vincent Pollet, and Mr. Bastien Crettol -- all members of Idiap staff. They contributed various software packages used in this work, helped with data-collection and other experiments.
%\vspace{-1mm}
%----------------------------------------------------------
\bibliographystyle{IEEEtran}
%\vspace{-1mm}
\bibliography{candy}

%----------------------------------------------------------

\end{document}